# Artifact Correction for Echo-Planar Imaging at Low-Field and Ultra-Low-Field MRI

Sisi Qiao, Yilin Yu, Tiecheng Lin, Yuhao Liu, Jiajia Sun, Xiaoling Li*

* Corresponding author at: School of Mechanical Engineering, Xi'an Jiaotong University, Xi'an, China

E-mail address: xjtulxl@mail.xjtu.edu.cn

## ARTICLEINFO

## ABSTRACT

Purpose: Echo-planar imaging (EPI) in low-field (LF) and ultra-low-field MRI (ULF) suffers from severe Nyquist ghost artifacts due to odd–even k-space misalignment. This study develops a reference-free artifact correction pipeline that reduces reliance on conventional reference scans while achieving improved ghost suppression.

Methods: Starting from the traditional reference-scan–based ghost artifact correction method, we first introduce a peak-alignment–based ghost artifact correction method to correct odd–even line displacement without reference data. To further reduce residual artifacts, an interpolation-and-resampling strategy is applied. The combined method was evaluated using EPI and diffusion-weighted EPI data in LF and ULF.

Results: The proposed pipeline effectively mitigated Nyquist ghosts, improved structural continuity, and enhanced signal uniformity. Peak-alignment–based ghost artifact correction method alone provided comparable artifact suppression to reference-scan–based ghost artifact correction method, while interpolation and resampling further suppressed residual artifacts, enabling reliable visualization of brain structures under ULF conditions.

Conclusion: A practical, reference-free correction pipeline is presented for LF and ULF EPI, combining peak-alignment–based ghost artifact correction method and interpolation-resampling to achieve efficient ghost suppression and expand the clinical applicability of low-field MRI systems, providing both theoretical guidance and practical experience for ULF EPI-based DWI imaging.

# 1. Introduction

Magnetic resonance imaging (MRI) continues to advance toward higher magnetic field strengths to support clinical and research applications. However, high-field systems are costly and largely confined to major medical centers, limiting accessibility. Recently, LF and ULF MRI have gained attention due to their cost-effectiveness, portability, and accessibility, despite inherently reduced signal-to-noise ratio (SNR) that limits image quality [1].

Stroke diagnosis relies on diffusion-weighted imaging (DWI), typically implemented with spin-echo (SE) sequences in high-field MRI (1.5–3 T). At LF and ULF, SE-based DWI suffers from extremely low SNR, producing noisy images. Extending scan times improves image quality but is impractical, motivating the use of EPI-based DWI sequences, which offer higher acquisition efficiency but introduce artifacts due to rapid gradient switching [2–4]. In our LF MRI prototype, EPI sequences outperformed SE-based methods in DWI performance.

Echo-planar imaging (EPI) [5] acquires the entire 2D k-space in a single excitation, providing relatively high SNR and enabling DWI [6,7], fMRI [8,9], and PWI [10]. Its high efficiency, however, introduces challenges in LF/ULF systems, particularly Nyquist ghost artifacts (1/2 ghosts) caused by misalignment between odd and even k-space lines [11]. These misalignments arise from echo reversal, long readouts, eddy currents, field inhomogeneity, susceptibility, and chemical shifts [12–15], manifesting as half-image ghosts that degrade image quality and quantitative accuracy.

Various strategies have been proposed for ghost correction. Reference scan–based methods [16–18], POCS [19], low-rank structured matrix approaches such as ALOHA [20], image-entropy minimization [21,22], and multi-coil GRAPPA [23,24] are effective in high-field MRI but require reference scans, multi-coil calibration, or complex computations. Deep learning methods [25–27] require large labeled datasets. In contrast, approaches analyzing k-space structural differences between odd and even lines [28,29] offer higher feasibility and lower computational complexity.

In this study, we address LF/ULF EPI ghost artifacts by progressing from reference scan–based correction to k-space–based odd–even line alignment. A peak-alignment–based ghost artifact correction method (method B) corrects misalignment without reference scans, reducing scan time while maintaining artifact suppression comparable to reference-scan–based ghost artifact correction method. To further reduce residual ghosts, an interpolation-and-resampling strategy (method C) enhances local line consistency. Together, these methods provide a lightweight, software-based solution for LF/ULF EPI, enabling effective DWI, improving image quality, and expanding clinical applicability.

# 2.Method

### A. Reference-scan-based ghost artifact correction method with phase-encoding turn-off

The principle of the reference scan is similar to that of a conventional EPI acquisition, except that the phase-encoding blip gradients are turned off and replaced by a pre-dephasing gradient. Each acquisition includes one reference scan echo and one corresponding imaging scan echo. By comparing the reference and imaging data, phase errors caused by B0 inhomogeneity, readout

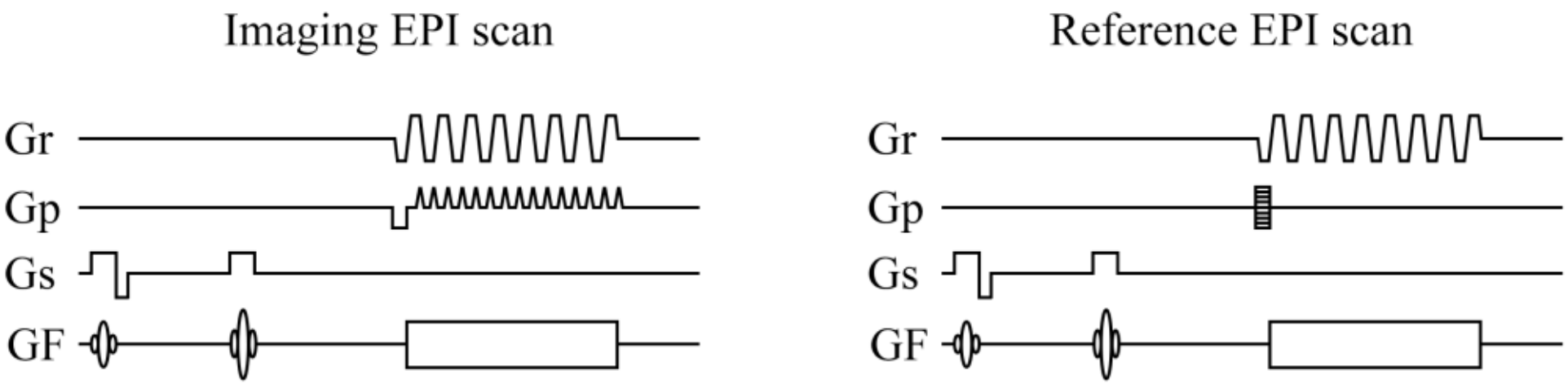

Fig. 1. Schematic of imaging EPI scan and reference EPI scan.

gradient–induced eddy currents, RF frequency offsets, and chemical shift effects can be estimated, while excluding eddy currents induced by the phase-encoding blip gradients [24]. Fig. 1 illustrates the sampling schemes of the imaging EPI scan and the reference EPI scan.

In EPI, misalignment between odd and even k-space lines results in N/2 (Nyquist) ghost artifacts. In the reference scan–based method, the phase difference between odd and even lines is extracted from the reference scan and applied to the corresponding imaging data for ghost correction.

Phase-based ghost correction can be performed using linear or nonlinear approaches. Both methods are effective when ghosting is dominated by echo peak shifts caused by timing imperfections. Under strong B0 inhomogeneity or off-resonance conditions, however, linear correction fails to account for echo distortion, whereas nonlinear correction may introduce streaking artifacts due to inaccurate phase estimation along the readout direction.

For each prescan acquisition, a reference echo corresponding to the subsequent imaging scan echo is collected. The estimated reference phase is mapped to the actual EPI data, enabling phase compensation between odd and even k-space lines and reducing Nyquist ghost artifacts.

In magnetic resonance imaging, the acquired raw data are complex-valued signals in k-space. Due to gradient polarity reversals in EPI acquisitions, line reversals are applied to ensure consistent orientation of odd and even lines. Each k-space sampling point contains a complex value represented by its real and imaginary components, as shown in Equation (2.1):

$$K(k_x, k_y) = Re(k_x, k_y) + i * Im(k_x, k_y) \tag{2.1}$$

Here, $Re(k_x, k_y)$ and $Im(k_x, k_y)$ denote the real and imaginary components of the signal, respectively.

The reference scan data $K_{ref}(k_x, k_y)$ are subjected to an inverse Fourier transform along the $k_x$ direction, as shown in Equation (2.2):

$$K_{ref}(x, k_y) = F^{-1}_{k_x \to x}\{K_{ref}(k_x, k_y)\} \tag{2.2}$$

Under LF and ULF conditions, the magnetic field is typically more stable near the center of the imaging volume but may exhibit local inhomogeneities toward the periphery. Therefore, the phase-angle error between odd and even lines is calculated by extracting the central adjacent odd and even lines from the reference scan data after the inverse Fourier transform, as shown in Equation (2.3):

$$\Delta\phi(x,ky) = atan2\left(\Im\{K_{ref33}(x,ky)\},\Re\{K_{ref33}(x,ky)\}\right) \\ -atan2\left(\Im\{K_{ref32}(x,ky)\},\Re\{K_{ref32}(x,ky)\}\right) \quad (2.3)$$

The actual imaging data $K_{formal}(k_x,k_y)$ are also subjected to an inverse Fourier transform along the $k_x$ direction, as shown in Equation (2.4):

$$K_{formal}(x,k_y) = F^{-1}_{k_x \to x}\{K_{formal}(k_x,k_y)\} \quad (2.4)$$

To perform phase-angle correction between odd and even lines in the imaging data, $K_{formal}(x,k_y)$ is first separated into odd-line data $K_{odd}$ and even-line data $K_{even}$. Their respective phase angles are computed as shown in Equations (2.5) and (2.6):

$$\phi_{odd}(x,k_y) = atan2\left(\Im\{K_{odd}(x,k_y)\},\Re\{K_{odd}(x,k_y)\}\right) \quad (2.5)$$
$$\phi_{even}(x,k_y) = atan2\left(\Im\{K_{even}(x,k_y)\},\Re\{K_{even}(x,k_y)\}\right) \quad (2.6)$$

The odd-line data (or alternatively the even-line data) are selected as the reference with their phase unchanged, while the even-line data are phase-corrected using the phase-angle error obtained from the reference scan, as shown in Equation (2.7):

$$\phi_{even_{cor}}(x,k_y) = \phi_{even}(x,k_y) + \Delta\phi(x,k_y) \quad (2.7)$$

Applying the corrected phase to the even-line magnitude data yields the corrected even-line signal, as shown in Equation (2.8):

$$K_{even_{cor}}(x,k_y) = \left|K_{even}(x,k_y)\right| * e^{i*\phi_{even_{cor}}(x,k_y)} \quad (2.8)$$

The fully corrected dataset is obtained by combining the phase-corrected even-line data with the unchanged odd-line data, as shown in Equation (2.9):

$$K_{cor}(x,k_y) = K_{even_{cor}}(x,k_y) + K_{odd}(x,k_y) \quad (2.9)$$

Performing a Fourier transform along the $k_x$ direction on the phase-corrected data $K_{cor}(x,k_y)$ yields the corrected k-space data $K_{cor}(k_x,k_y)$, as shown in Equation (2.10) or performing an inverse Fourier transform along the $k_y$ direction on $K_{cor}(x,k_y)$ produces the corrected spatial-domain data $K_{cor}(x,y)$, as shown in Equation (2.11):

$$K_{cor}(k_x,k_y) = F_{x \to k_x}\{K_{formal}(x,k_y)\} \quad (2.10)$$

$$K_{cor}(x,y) = F^{-1}_{k_y \to y}\{K_{formal}(x,k_y)\} \quad (2.11)$$

**B. Peak-alignment–based ghost artifact correction method**

The reference scan method inevitably doubles the acquisition time for a single EPI scan. Its core objective is to identify and compensate for the misalignment between odd and even k-space lines responsible for Nyquist ghost artifacts. While the reference scan enables direct estimation of phase-angle errors in the (x, $k_y$) domain, accurate phase extraction from actual scan data without a reference scan is challenging due to phase-encoding blips, B0 inhomogeneity, and the presence of true imaging phase information.

Experimental observations indicate that in the original ($k_x$, $k_y$) k-space, the relative peak positions of odd and even lines are strongly correlated with Nyquist ghost severity. After reference scan correction, these peaks tend to align. Motivated by this observation, a reference-free peak-alignment–based ghost artifact correction method is proposed, in which Nyquist ghosting is mitigated by directly aligning the peak positions of odd and even k-space lines.

The peak-alignment–based ghost artifact correction method operates directly on the original k-space data. First, the acquired data are divided into odd and even line subsets according to the parity of the row indices, as shown in Equation (2.12):

$$\begin{cases} K_{odd}(k_x, k_y),\ k_y = odd\ numbers \\ K_{even}(k_x, k_y),\ k_y = even\ numbers \end{cases} \tag{2.12}$$

For each k-space row $K_i$, the position of the maximum signal magnitude is defined as shown in Equation (2.13):

$$p_i = \arg max|K_i(k_x, k_y)| \tag{2.13}$$

The central k-space rows—rows 32 and 33—are selected as reference rows for the even and odd data, respectively. Their peak positions are calculated as shown in Equations (2.14) and (2.15):

$$p_{32} = \arg max|K_{32}(k_x, k_y)| \tag{2.14}$$

$$p_{33} = \arg max|K_{33}(k_x, k_y)| \tag{2.15}$$

The relative peak shift $\Delta p$ between odd and even lines is then calculated as the difference between the peak positions of the two reference rows, as shown in Equation (2.16):

$$\Delta p = p_{33} - p_{32} \tag{2.16}$$

All even lines (or alternatively all odd lines) are shifted by the corresponding $\Delta p$ along the $k_x$ direction to align the peak positions between odd and even k-space lines. The shifted even-line data are denoted as $\widetilde{K_{even}}(k_x, k_y)$, as shown in Equation (2.17):

$$\widetilde{K_{even}}(k_x, k_y) = K_{even}(k_x + \Delta p, k_y) \tag{2.17}$$

The final peak-alignment–corrected k-space data $K_{cor}(k_x, k_y)$ are obtained by recombining

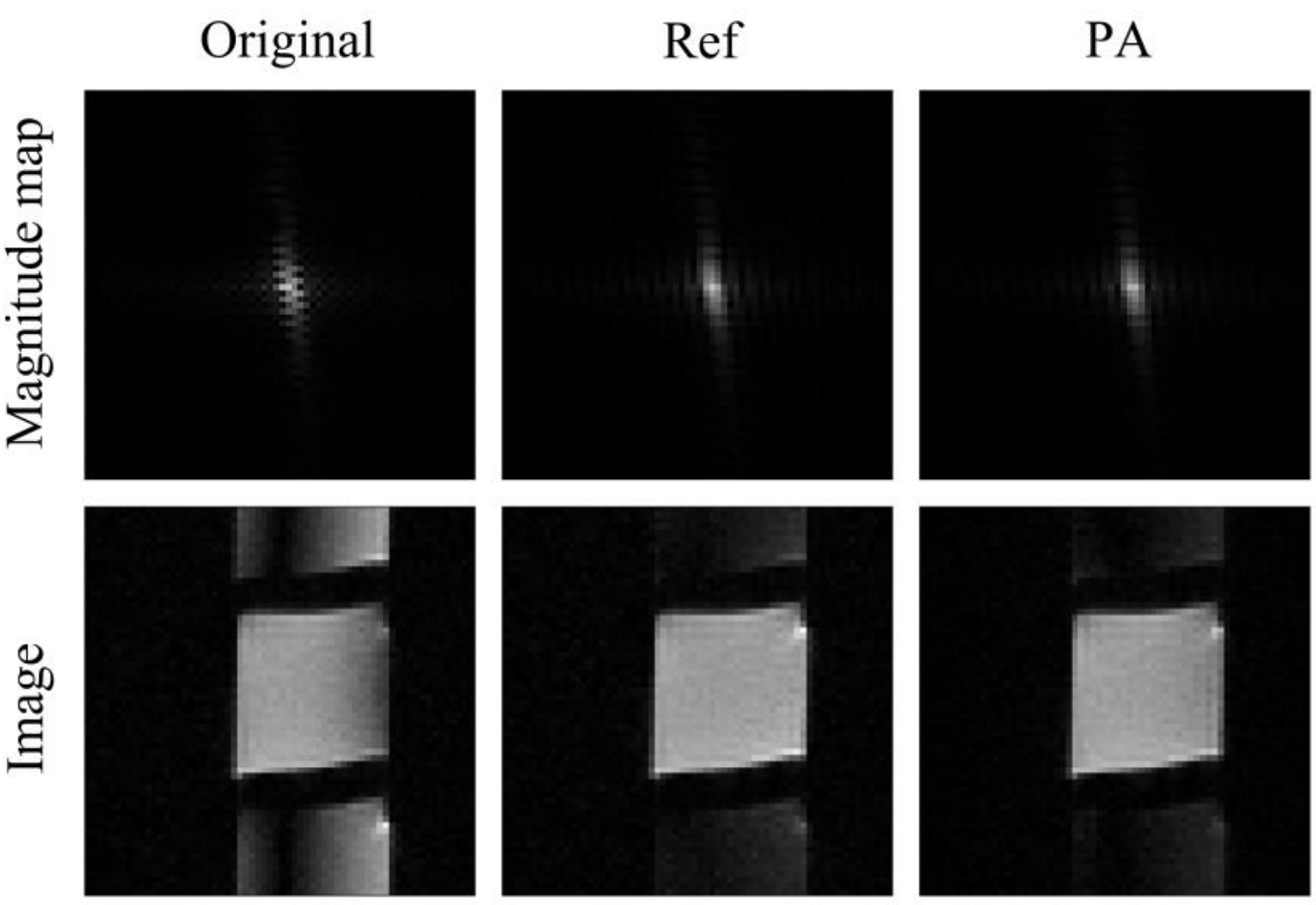

Fig. 2. K-space magnitude maps and water phantom images before and after correction. The figure consists of three columns: Original, showing the uncorrected images; Ref, showing images corrected using Method A, reference-scan–based ghost artifact correction with phase-encoding turned off; and PA, showing images corrected using Method B, peak-alignment–based ghost artifact correction.

the original odd lines with the shifted even lines, as shown in Equation (2.18):

$$K_{cor}(k_x, k_y) = K_{odd}(k_x, k_y) + \widetilde{K_{even}}(k_x, k_y) \tag{2.18}$$

Through this peak-alignment operation in k-space, the dominant misalignment between odd and even lines is effectively corrected, resulting in substantial suppression of Nyquist ghost artifacts. Similar to the reference scan–based method, residual ghost artifacts may still remain after peak alignment. Therefore, additional residual ghost correction steps are subsequently applied to obtain the final corrected data.

**C. Interpolation-and-resampling-based residual ghost artifact correction method**

After applying reference scan–based correction (Method A) or peak-alignment–based ghost artifact correction method (Method B), both original and corrected datasets are reconstructed in the k-space and spatial domains. K-space magnitude images are used to assess line-wise consistency, while spatial-domain images evaluate overall image quality. Fig. 2 compares k-space magnitude maps and water phantom images before and after applying Methods A and B.

Although both methods effectively suppress dominant odd–even inconsistencies, residual ghost artifacts remain visible. Fig. 3 shows pronounced left–right asymmetry, primarily caused by insufficient magnetic field homogeneity in LF and ULF MRI systems. Rapid gradient reversals further exacerbate local mismatches along each readout line, leading to residual ghost artifacts.

Another contributing factor is the limited sampling resolution, typically restricted to a 64 × 64 grid in LF and ULF EPI. Undersampling introduces residual inconsistencies between odd and even lines, indicating that perfect alignment after preliminary correction may not be achievable.

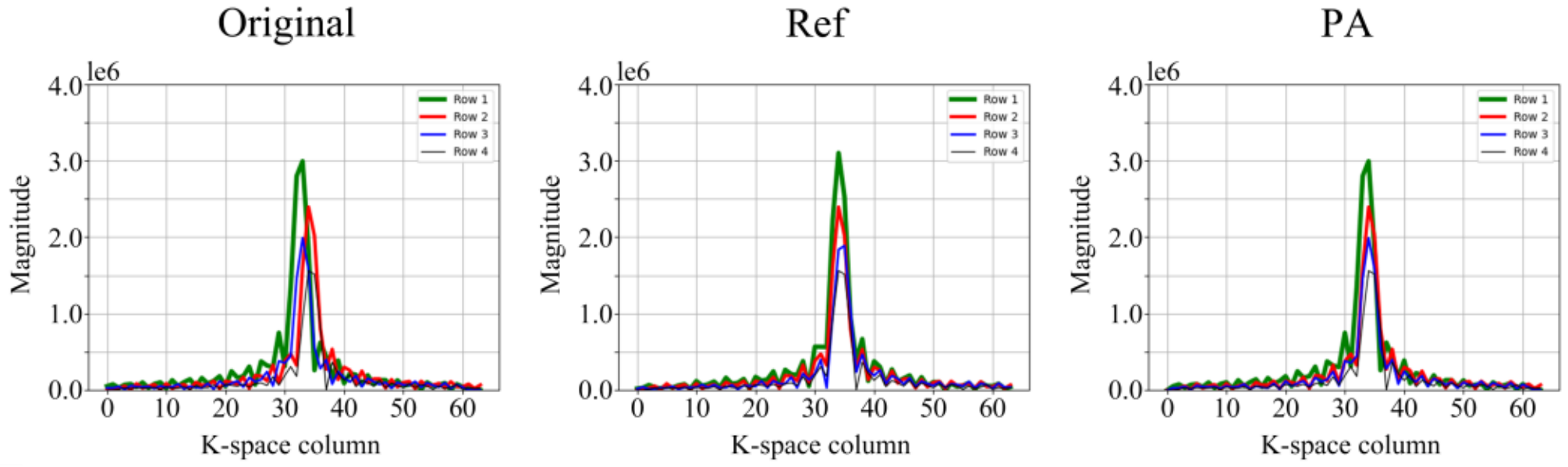

Fig. 3. Row-wise magnitude profiles of the original data and the data corrected using the Ref and PA methods. The figure consists of three columns: Original, showing the profiles extracted from uncorrected data; Ref, showing the profiles after correction using Method A, the reference-scan–based ghost artifact correction with phase-encoding turned off; and PA, showing the profiles after correction using Method B, the peak-alignment–based ghost artifact correction.

To address residual ghost artifacts that cannot be fully corrected by Methods A or B, an interpolation-and-resampling (IR) strategy is introduced to further reduce local inconsistencies between odd and even lines. The interpolation is applied to the spatial frequency–domain data after preliminary ghost correction.

The data before interpolation are denoted as $K_{cor}(k_x, k_y) \in C^{N\times M}$, where N and M represent the number of columns and rows, respectively. The interpolation factor is denoted as shown in Equation (2.19):

$$f = \frac{N_{interp}}{N} \tag{2.19}$$

Here $N_{interp}$ denotes the number of columns after interpolation. The interpolated data are denoted as $\widetilde{K_{cor}}(k_x, k_y) \in C^{(f*N)\times M}$. Subsequently, the interpolated data are then resampled to restore the original data structure, yielding $\widehat{K_{cor}}(k_x, k_y) \in C^{N\times M}$.

For convenience of notation, the discrete representations are defined as follows: the corrected k-space data $K_{cor}(k_x, k_y)$ are denoted as $K_{cor}[n, m]$, where $n = 0, \ldots, N-1$ corresponds to $k_x$ and $m = 0, \ldots, M-1$ corresponds to $k_y$; the interpolated data $\widetilde{K_{cor}}(k_x, k_y)$ are denoted as $\widetilde{K_{cor}}[p, m]$, where $p = 0, \ldots, f*N-1$ corresponds to $k_x$ and $m = 0, \ldots, M-1$ corresponds to $k_y$; the resampled data $\widehat{K_{cor}}(k_x, k_y)$ are denoted as $\widehat{K_{cor}}[n, m]$, where $n = 0, \ldots, N-1$ corresponds to $k_x$ and $m = 0, \ldots, M-1$ corresponds to $k_y$.

Interpolation is performed column by column along the $k_x$ direction. For a given column $k_y = m$, the original column vector is defined as $k^{(m)}[n] = K[n, m]$.

Let $I_f$ denote the interpolation operator. The interpolated vector is obtained as shown in Equation (2.20):

$$\tilde{k}^{(m)}[p] = I_f\{k^{(m)}\}[p], \quad p = 0, \ldots, f*N-1 \tag{2.20}$$

Using discrete linear interpolation, the index $p$ can be expressed as $p = n*f + r$, where $n = \lfloor p/f \rfloor$ is the integer quotient and $r = p \bmod f$ is the remainder. Accordingly, the interpolated

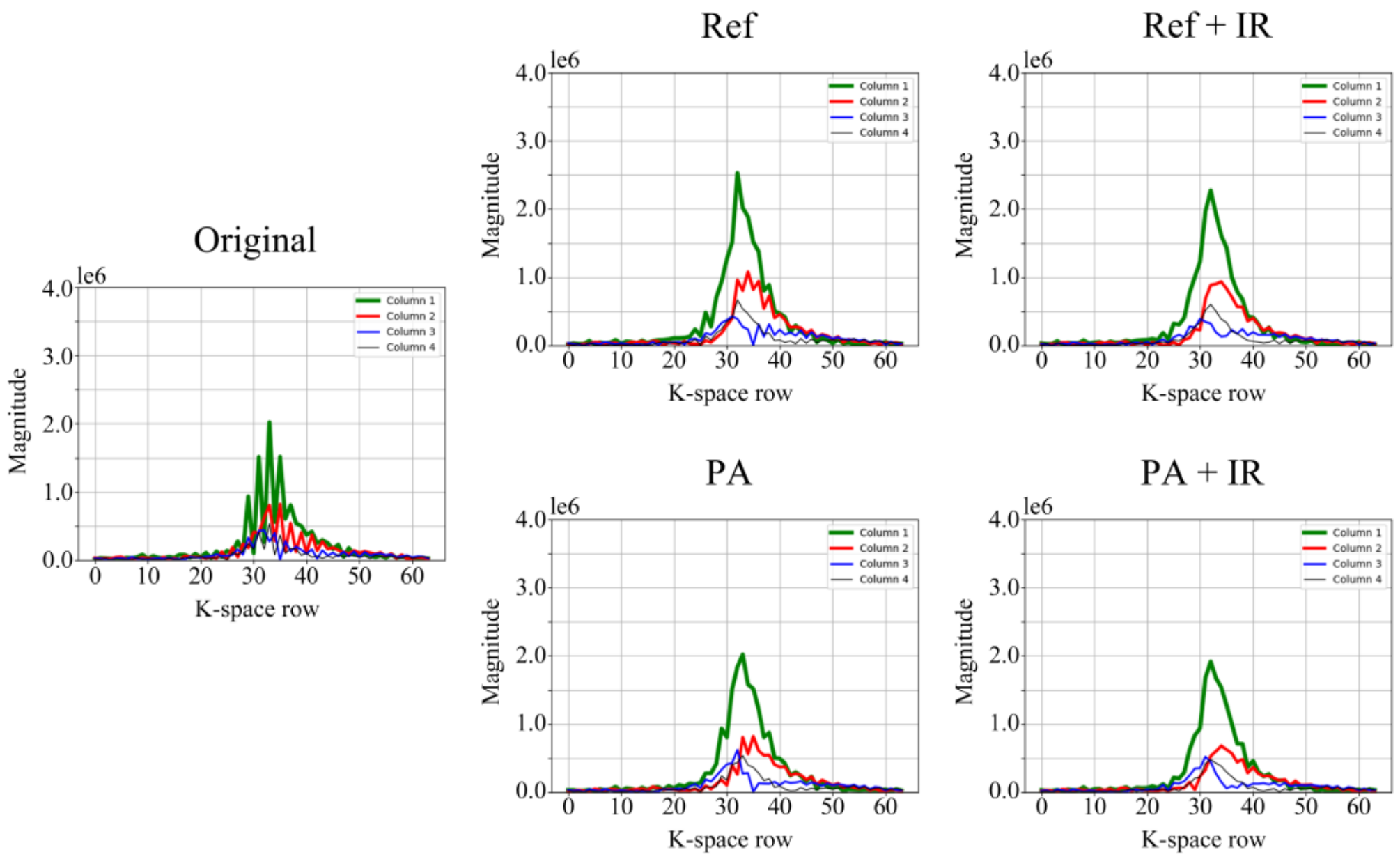

Fig. 4. Column-wise magnitude profiles of the original data and the data corrected using the Ref, Ref + IR, PA, and PA + IR methods. The figure consists of five columns: Original, showing the profiles extracted from uncorrected data; Ref, showing the profiles after correction using Method A, the reference-scan–based ghost artifact correction with phase-encoding turned off; Ref + IR, showing the profiles after sequential correction using Method A followed by Method C, the interpolation-and-resampling–based residual ghost artifact correction; PA, showing the profiles after correction using Method B, the peak-alignment–based ghost artifact correction; and PA + IR, showing the profiles after sequential correction using Method B followed by Method C.

values are computed as shown in Equation (2.21):

$$\tilde{k}^{(m)}[p] = \begin{cases} k^{(m)}[n], r = 0 \\ (1-\alpha)k^{(m)}[n] + \alpha k^{(m)}[n+1], r > 0, \end{cases} \alpha = \frac{r}{f} \tag{2.21}$$

In matrix form, the interpolation process is written as: $\widetilde{K_{cor}} = I_f(K_{cor})$, $\widetilde{K_{cor}} \in C^{(f*N)\times M}$.

The resampling operator after interpolation is defined as $R_f$, which extracts every f-th point at uniform intervals to restore the original column resolution. The resampled vector is obtained as shown in Equation (2.22):

$$\hat{k}^{(m)}[n] = R_f\{\tilde{k}^{(m)}\}[n] = \tilde{k}^{(m)}[n*f], \quad n = 0, \ldots, N-1 \tag{2.22}$$

In matrix form, the restored data are expressed as: $\widehat{K_{cor}} = R_f(\widetilde{K_{cor}})$, $\widehat{K_{cor}} \in C^{N\times M}$.

The interpolation parameter $N_{interp}$, representing the number of points after expansion, was tested using values of 128, 256, 512, and higher. Multiple rounds of interpolation and resampling were performed to identify the optimal correction parameter for further suppression of residual ghost artifacts. Experimental results demonstrated that the most effective artifact reduction was achieved

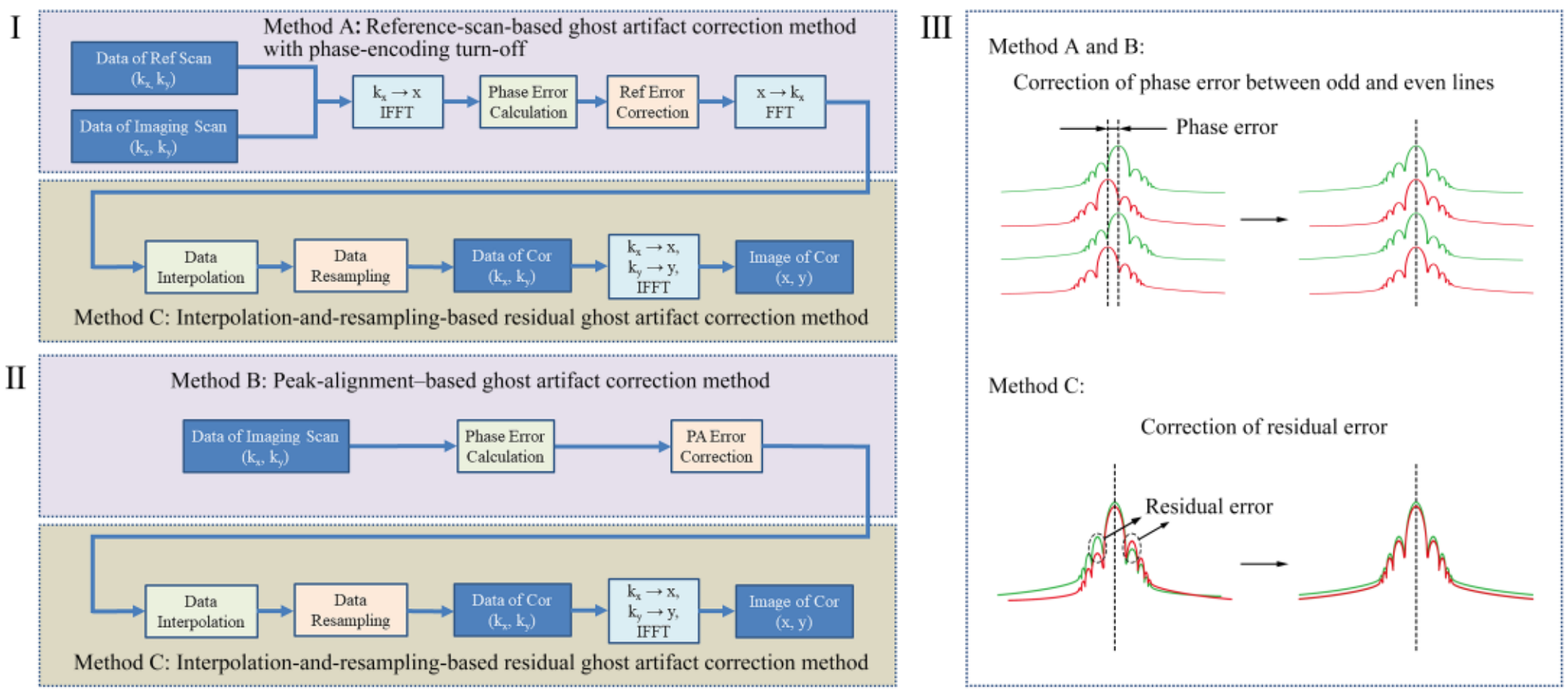

Fig. 5. Flowchart of EPI ghost artifact correction and schematic illustration of the data correction procedures.

I: Correction pipeline combining A. Reference-scan–based ghost artifact correction with phase-encoding turned off and C. Interpolation-and-resampling–based residual ghost artifact correction.

II: Correction pipeline combining B. Peak-alignment–based ghost artifact correction and C. Interpolation-and-resampling–based residual ghost artifact correction.

III: Schematic illustration of the preliminary ghost artifact correction achieved by methods A and B, and the residual ghost artifact correction performed by method C.

when the number of interpolation points was approximately 4096. Therefore, a 64× interpolation factor was selected as the correction parameter in subsequent processing.

The column-wise magnitude profiles before and after interpolation and resampling are shown in Fig. 4. After applying the IR method, the profiles become noticeably smoother, indicating improved local consistency between adjacent rows. The reduction in inter-row fluctuations corresponds to a decrease in residual ghost artifacts in the spatial domain, thereby indirectly validating the theoretical effectiveness of the interpolation-and-resampling–based residual ghost artifact correction method.

### D. Integrated EPI ghost artifact correction framework

The overall workflow of Methods A, B, and C, together with their corresponding data correction procedures, is summarized in Fig. 5. Both the reference-scan–based ghost artifact correction method (Method A) and the reference-free peak-alignment–based ghost artifact correction method (Method B) serve as preliminary EPI ghost artifact correction steps, targeting the dominant odd–even line inconsistencies. Method C, the interpolation-and-resampling–based residual ghost artifact correction method, is subsequently applied to further suppress residual ghost artifacts when combined with either Method A or Method B.

Based on these components, an integrated EPI ghost artifact correction framework was established for LF and ULF MRI. The framework consists of two complementary correction pipelines: (I) a reference-scan–based pipeline that combines Method A with Method C, and (II) a reference-free pipeline that combines Method B with Method C. This design enables flexible selection of the correction strategy depending on the availability of reference scans, while ensuring consistent suppression of both dominant and residual ghost artifacts.

For clarity and consistency in figure annotations throughout this study, the original uncorrected images are denoted as Original. The reference-scan–based ghost artifact correction method with phase encoding turned off (Method A) is abbreviated as Ref, the peak-alignment–based ghost artifact correction method (Method B) is abbreviated as PA, and the interpolation-and-resampling–based residual ghost artifact correction method (Method C) is abbreviated as IR.

## 3.Experimental

EPI scan data were acquired using experimental MRI prototype systems operating at magnetic field strengths of 0.5 T and 0.068 T. All image reconstruction procedures and EPI ghost artifact correction methods evaluated in this study were implemented in Python. Imaging experiments were performed on water phantoms and healthy volunteers to evaluate the performance of the proposed EPI ghost artifact correction methods.

The EPI acquisition parameters for the 0.5 T MRI prototype were as follows: matrix size = 64 × 64, field of view (FOV) = 250 mm × 250 mm, slice thickness = 5 mm, echo time (TE) = 86 ms, repetition time (TR) = 3000 ms, and number of averages = 1.

The EPI acquisition parameters for the 0.068 T MRI prototype were as follows: matrix size = 64 × 64, field of view (FOV) = 350 mm × 350 mm, slice thickness = 20 mm, echo time (TE) = 171 ms, repetition time (TR) = 6000 ms, and number of averages = 4.

The DW-EPI acquisition parameters for the 0.068 T MRI prototype were as follows: matrix size = 64 × 64, field of view (FOV) = 350 mm × 350 mm, slice thickness = 20 mm, echo time (TE) = 171 ms, repetition time (TR) = 6000 ms, number of averages = 4, and b-value = 0 or 400 s/mm$^2$.

## 4.Results and discussion

Based on the reference-scan–based ghost artifact correction method, the peak-alignment–based ghost artifact correction method, and the interpolation-and-resampling–based residual ghost artifact correction method described above, the results of the water phantom experiments are shown in Fig. 6. Compared with the uncorrected images, both the reference-scan–based ghost artifact correction method and the peak-alignment–based ghost artifact correction method achieve an initial reduction of ghost artifacts in low-field MRI images. Building on these corrections, the subsequent application of the interpolation-and-resampling–based residual ghost artifact correction method further suppresses the remaining ghost artifacts and improves overall image quality.

By incorporating the interpolation-and-resampling–based residual ghost artifact correction method, image artifacts are further reduced beyond the preliminary correction achieved by the other two methods, demonstrating the effectiveness and robustness of the proposed residual ghost artifact correction strategy. Compared with the reference-scan–based ghost artifact correction method, the reference-free peak-alignment–based ghost artifact correction method achieves comparable preliminary ghost artifact reduction while eliminating the need for an additional reference scan. For the current system and sequence, the reference scan requires an acquisition time equal to that of the actual scan; therefore, the reference-free approach reduces the total scan time by approximately 50%, which is highly advantageous for clinical applications.

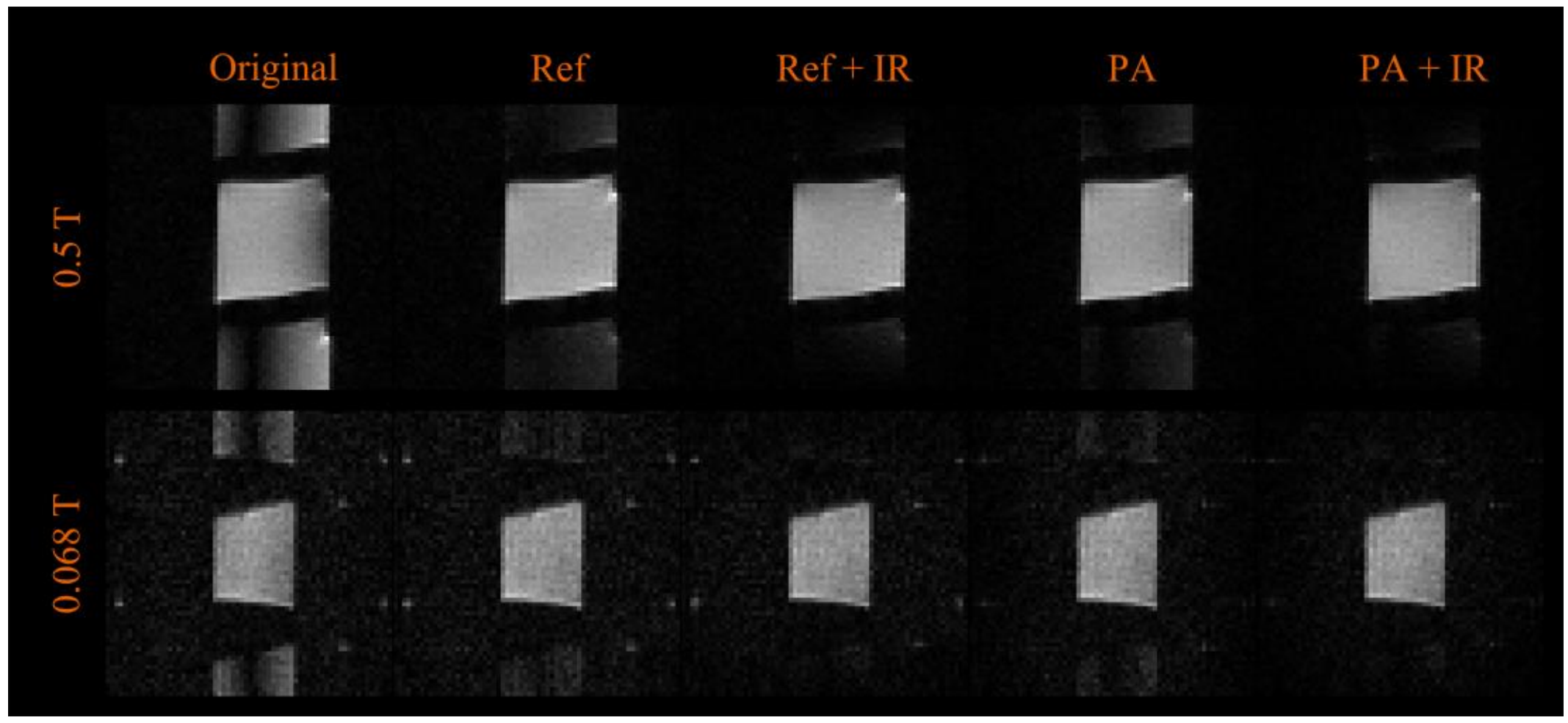

Fig. 6. Comparison of EPI ghost correction results on 0.5 T and 0.068 T MRI systems. Five columns of images show water phantom data corresponding to the Original, Ref, Ref + IR, PA, and PA + IR methods. Two rows of images represent the results acquired on the 0.5 T and 0.068 T MRI systems, respectively.

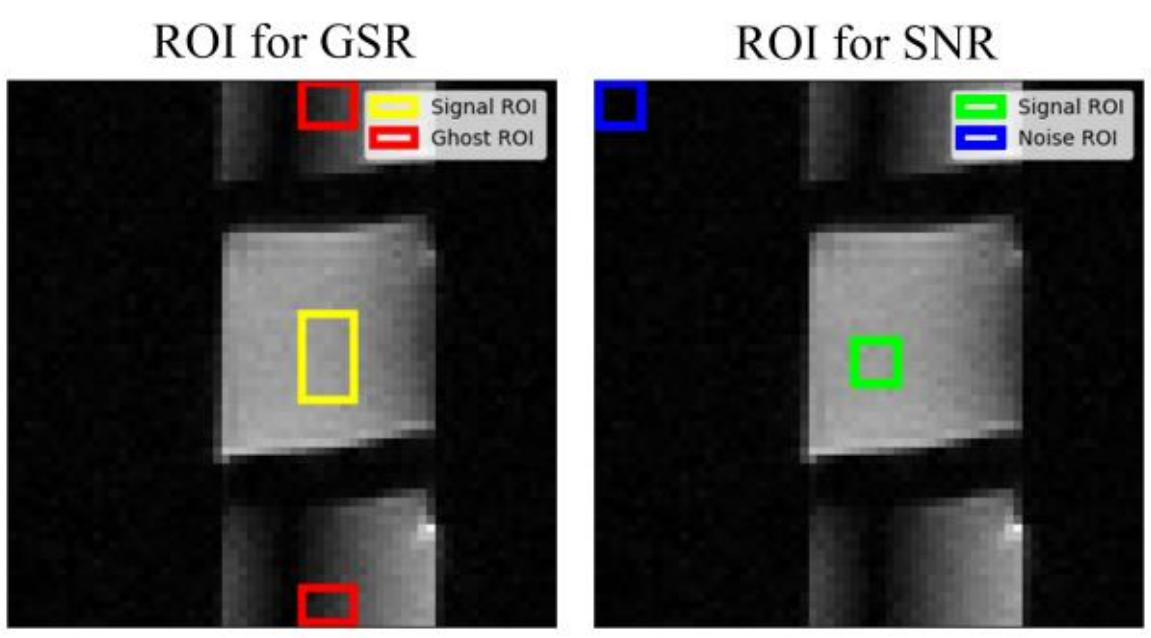

Fig. 7. Illustration of the ROIs used for GSR and SNR calculation.

The objective evaluation of the above methods was conducted using the ghost-to-signal ratio (GSR) [30,31] and the signal-to-noise ratio (SNR) [32,33]. The regions of interest (ROIs) used for GSR and SNR evaluation are illustrated in Fig. 7.

To quantitatively assess ghost artifact suppression, the GSR was calculated for images before and after correction. An ROI in the central region of the image was defined as the true image region, and a corresponding ghost artifact region was obtained by shifting this ROI by N/2 pixels along the y direction, according to the N/2 ghost distribution property. This ROI selection reflects the fact that, in low-field MRI, diagnostically relevant structures are primarily located in the central region, whereas peripheral regions are more susceptible to magnetic field inhomogeneity. The GSR was defined as the ratio of the mean signal intensity in the ghost artifact region to that in the true image region. The quantitative results are summarized in Table 1.

For the water phantom experiment shown in Fig. 6, on the 0.5 T system, the GSR of the uncorrected image was 0.171. Application of the reference-scan–based ghost artifact correction method reduced the GSR to 0.097, while the peak-alignment–based ghost artifact correction method achieved a further reduction to 0.041. When combined with the 64-fold single-pass linear interpolation-and-resampling–based residual ghost artifact correction method, the GSR was further reduced to 0.022 for the reference-scan–based ghost artifact correction method and to 0.017 for the peak-alignment–based ghost artifact correction method. On the 0.068 T system, the GSR decreased from 0.253 (uncorrected) to 0.179 and 0.108 after applying the reference-scan–based ghost artifact correction method and peak-alignment–based ghost artifact correction method alone, respectively, and was further reduced to 0.054 and 0.045 after incorporating the interpolation-and-resampling–based residual ghost artifact correction method.

Table 1. GSR and SNR before and after EPI ghost correction on 0.5 T and 0.068 T MRI systems

| Condition | Metric | Method | | | | |
|---|---|---|---|---|---|---|
| | | Original | Ref | Ref + IR | PA | PA + IR |
| Phantom at 0.5 T | GSR | 0.1710 | 0.0973 | 0.0218 | 0.0412 | 0.0172 |
| | SNR | 43.74 | 39.17 | 82.47 | 74.23 | 131.93 |
| Phantom at 0.068 T | GSR | 0.2525 | 0.1795 | 0.0535 | 0.1079 | 0.0454 |
| | SNR | 19.72 | 19.45 | 31.35 | 35.64 | 70.17 |

Overall, GSR-based quantitative evaluation demonstrates that the peak-alignment–based ghost artifact correction method provides superior preliminary ghost suppression compared with the reference-scan–based ghost artifact correction method. Moreover, the interpolation-and-resampling–based residual ghost artifact correction method consistently enables additional reduction of residual ghost artifacts. Comparison between the two systems further indicates that EPI images acquired at lower magnetic field strengths suffer from more severe ghost artifacts, underscoring the importance of effective ghost artifact correction in low-field MRI.

Using the signal-to-noise ratio (SNR) as an additional quantitative metric, an ROI in the central region of the image was defined as the true image region, while an ROI in the image corner was selected as the noise region. The SNR was calculated as the ratio of the mean signal intensity in the central region to that in the noise region. The quantitative results are listed in Table 1.

For the water phantom experiment shown in Fig. 6, on the 0.5 T system, the SNR of the uncorrected image was 43.7. Application of the reference-scan–based ghost artifact correction method reduced the SNR to 39.2, whereas the peak-alignment–based ghost artifact correction method increased the SNR to 74.2. When combined with the 64-fold single-pass linear interpolation-and-resampling–based residual ghost artifact correction method, the SNR further increased to 82.5 for the reference-scan–based ghost artifact correction method and to 131.9 for the peak-alignment–based ghost artifact correction method. On the 0.068 T system, the SNR increased from 19.7 (uncorrected) to 35.6 after applying the peak-alignment–based ghost artifact correction method alone, while a slight decrease to 19.5 was observed for the reference-scan–based ghost artifact correction method. Further incorporation of the interpolation-and-resampling–based residual ghost artifact correction method increased the SNR to 31.3 and 70.2 for the reference-scan–based ghost artifact correction method and peak-alignment–based ghost artifact correction method, respectively.

Overall, SNR-based quantitative evaluation indicates that the reference-scan–based ghost artifact correction method introduces additional noise from the reference scan, leading to a reduction in SNR during preliminary correction. In contrast, the peak-alignment–based ghost artifact correction method and the interpolation-and-resampling–based residual ghost artifact correction method improve data consistency and effectively suppress ghost artifacts, resulting in increased

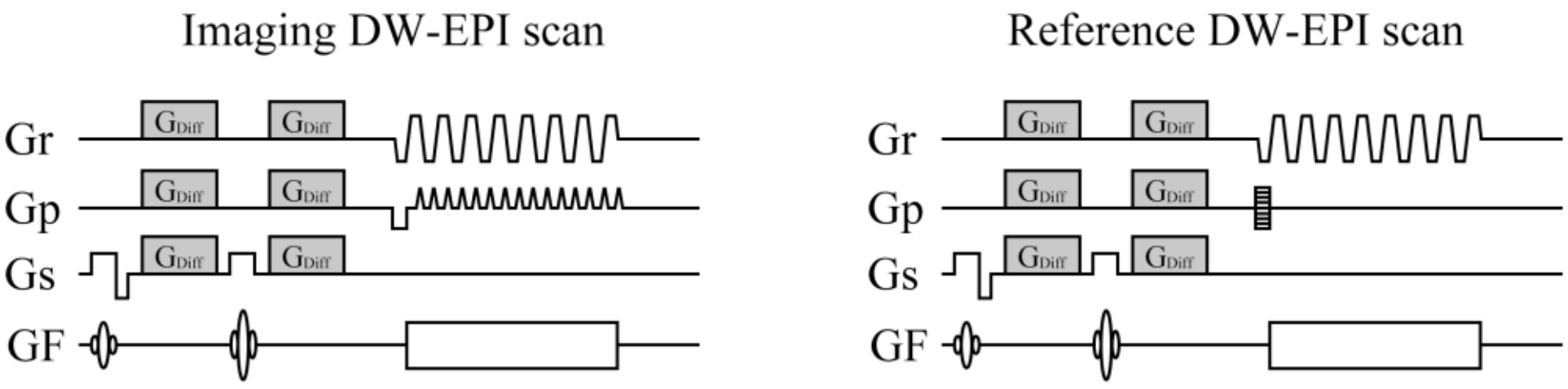

Fig. 8. Schematic of Imaging DW-EPI scan and Reference DW-EPI scan

SNR. Consequently, the peak-alignment–based ghost artifact correction method demonstrates superior SNR performance during the preliminary correction stage, and its combination with the interpolation-and-resampling–based residual ghost artifact correction method further enhances overall image quality.

The water phantom experiments conducted on both systems, evaluated using qualitative image assessment as well as quantitative GSR and SNR metrics, confirm the effectiveness of the proposed algorithms across different devices and imaging conditions. Comparative results indicate that the peak-alignment–based ghost artifact correction method combined with the 64-fold single-pass linear interpolation-and-resampling–based residual ghost artifact correction method provides slightly better Nyquist ghost suppression than the corresponding reference-scan–based ghost artifact correction method in both LF and ULF MRI.

Following effective correction of conventional EPI ghost artifacts, the proposed algorithm was further evaluated on diffusion-weighted EPI (DW-EPI). The sampling schemes of the DW-EPI imaging scan and the corresponding reference scan are illustrated in Fig. 8. Due to the inclusion of diffusion-weighting gradients, DW-EPI exhibits reduced SNR and more severe geometric distortions compared with conventional EPI, posing additional challenges for ghost artifact correction. To assess algorithm robustness under these conditions, DW-EPI experiments were performed on a 0.068 T MRI system using a water phantom and healthy volunteers. Representative results are shown in Fig. 9.

Both phantom and in vivo experiments included DW-EPI acquisitions with b-values of 0 and 400 s/mm². Visual inspection demonstrates that the proposed artifact correction algorithm effectively suppresses ghost artifacts under both diffusion-weighting conditions. Quantitative evaluation using GSR and SNR further confirms these observations. For both phantom and volunteer experiments at 0.068 T, the peak-alignment–based ghost artifact correction method combined with interpolation-and-resampling–based residual ghost artifact correction method achieved the best overall performance in terms of both metrics. In contrast, the reference-scan–based ghost artifact correction method alone did not yield noticeable improvements and, in some cases, resulted in degraded performance, indicating that diffusion-weighting further aggravates image noise and may cause noise amplification in reference-scan–based ghost artifact correction.

Similarly, the peak-alignment–based ghost artifact correction method alone provided limited improvement in GSR and SNR under the increased noise level of DW-EPI. These results highlight the critical role of the interpolation-and-resampling–based residual ghost artifact correction step under ultra-low-field conditions. By incorporating this final correction, ghost artifacts are substantially suppressed while noise is simultaneously reduced, resulting in improved image quality. The quantitative results are summarized in Table 2.

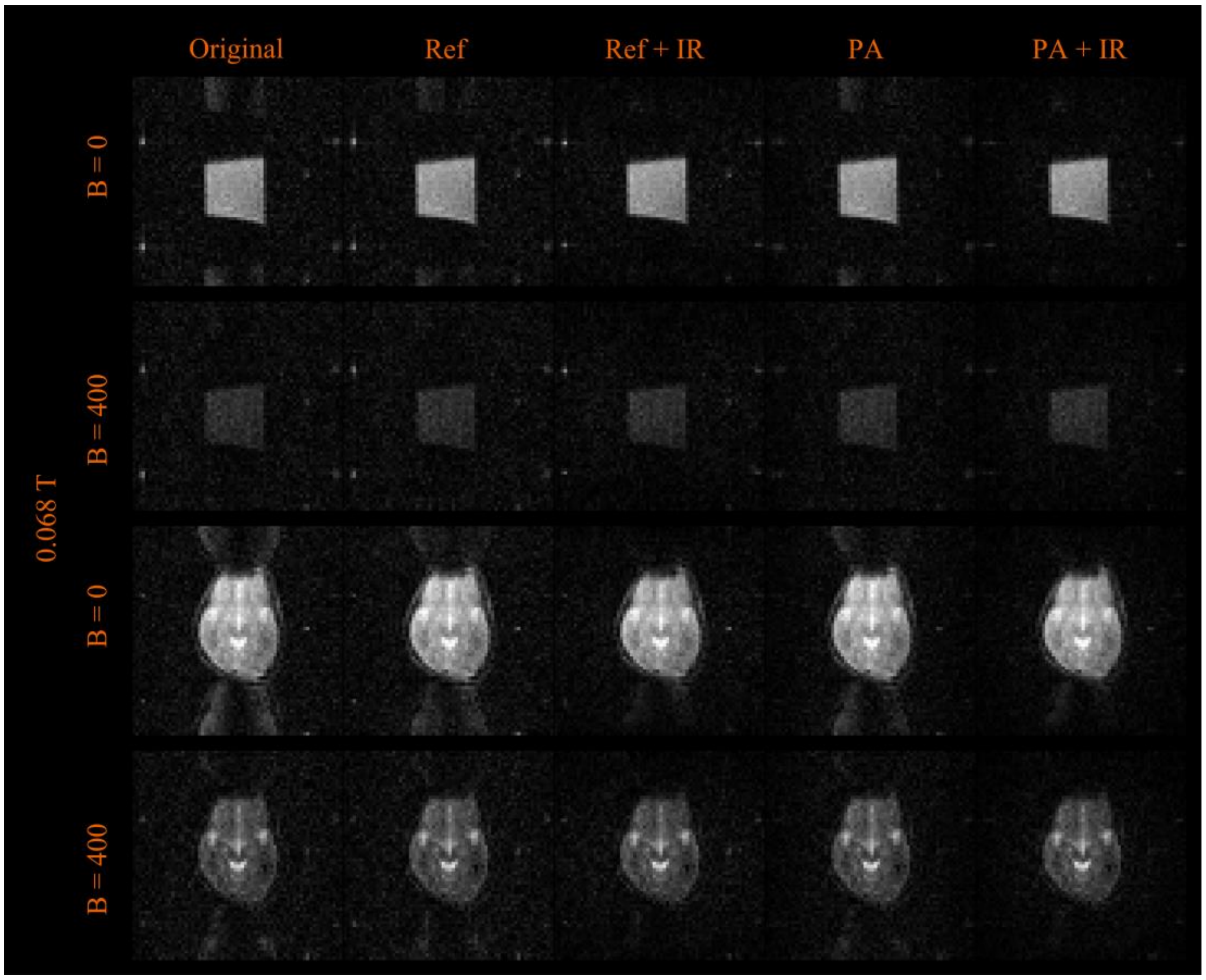

Fig. 9. Comparison of DW-EPI ghost correction results on the 0.068 T MRI system. Five columns of images show data corresponding to the Original, Ref, Ref + IR, PA, and PA + IR methods. The top two rows show water phantom images acquired with b = 0 and b = 400, respectively. The bottom two rows show volunteer images acquired with b = 0 and b = 400, respectively.

Subsequently, seven independent experiments were conducted to statistically evaluate the artifact correction performance of the peak-alignment–based ghost artifact correction method combined with interpolation-and-resampling–based residual ghost artifact correction method, including five water phantom experiments and two volunteer experiments. Each experiment comprised two DW-EPI acquisitions with b-values of 0 and 400 s/mm², both processed using the proposed correction algorithm. The statistical results are summarized in Fig. 10, which shows the percentage of residual artifacts relative to the uncorrected images.

Under the current experimental conditions, the peak-alignment–based ghost artifact correction method combined with interpolation-and-resampling–based residual ghost artifact correction method reduced the artifact level to approximately 30%–50% of the original level. For acquisitions with b = 400 s/mm², the uncorrected images exhibited higher initial artifact levels than those acquired with b = 0 s/mm², resulting in a larger relative artifact reduction after correction

Table 2. GSR and SNR before and after DW-EPI ghost correction on 0.068 T MRI systems

| Condition | Metric | Method | | | | |
|---|---|---|---|---|---|---|
| | | Original | Ref | Ref + IR | PA | PA + IR |
| Phantom at b = 0 s/mm² | GSR | 0.1288 | 0.1350 | 0.0478 | 0.1132 | 0.0460 |
| | SNR | 17.78 | 17.15 | 30.26 | 28.88 | 52.95 |
| Phantom at b = 400 s/mm² | GSR | 0.2397 | 0.2727 | 0.1438 | 0.2296 | 0.1304 |
| | SNR | 5.30 | 5.27 | 10.42 | 10.13 | 19.84 |
| Volunteer at b = 0 s/mm² | GSR | 0.1063 | 0.1328 | 0.0450 | 0.0870 | 0.0388 |
| | SNR | 23.04 | 19.04 | 32.26 | 31.07 | 66.11 |
| Volunteer at b = 400 s/mm² | GSR | 0.1343 | 0.1286 | 0.0636 | 0.1183 | 0.0618 |
| | SNR | 11.34 | 11.36 | 26.54 | 24.71 | 46.58 |

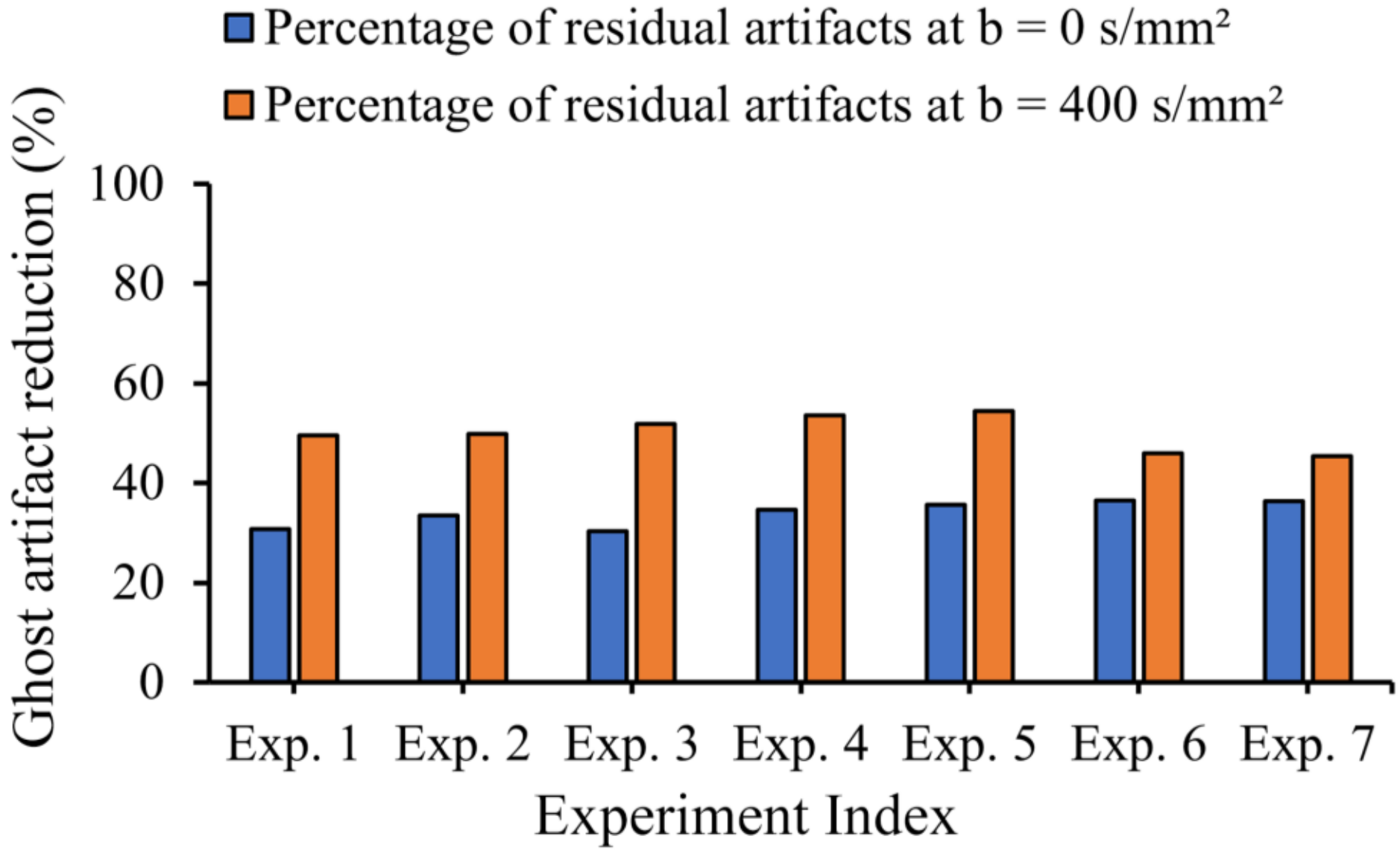

Fig. 10. Statistical evaluation of the PA + IR method for DW-EPI ghost artifact correction. Seven independent experiments were conducted: Experiments 1–5 were performed on phantoms, and Experiments 6–7 were conducted on volunteers. Each experiment included acquisitions with b = 0 and b = 400 s/mm². The figure shows the percentage of residual artifacts relative to the original uncorrected images after correction.

## 5.Conclusion

This study extends existing EPI artifact correction methods originally developed for high-field MRI to LF and ULF MRI systems. By visualizing k-space data, the underlying causes of residual ghost artifacts were identified, motivating the development of an interpolation-and-resampling–based residual artifact correction strategy. In addition, an odd–even peak alignment strategy was proposed by analyzing peak position misalignment between odd and even k-space lines before and after reference-based correction. When combined with interpolation-based resampling, this strategy enables accurate odd–even line alignment without requiring reference scans, thereby reducing acquisition time compared with conventional reference-scan–based ghost artifact correction method. Experimental results demonstrate that the proposed approach effectively suppresses Nyquist ghost artifacts in EPI imaging for LF and ULF MRI, leading to improved SNR. Furthermore, the method was successfully applied to DW-EPI sequences, substantially enhancing diffusion-weighted imaging quality at low and ultra-low magnetic field strengths.

### A. Effectiveness of the Correction Method

The proposed odd–even peak alignment–based artifact correction method operates directly on k-space data and achieves relative alignment between odd and even lines by matching magnitude peaks. Experimental analysis of row-wise magnitude profiles shows that peak translation effectively aligns odd and even data, corresponding to an initial suppression of ghost artifacts in the reconstructed images. Unlike conventional approaches that rely on additional reference scans or complex matrix decomposition, the proposed method requires only the acquired EPI data to achieve preliminary artifact correction, thereby reducing acquisition time and improving practical applicability.

Building upon the initial peak alignment, the interpolation-and-resampling–based residual correction addresses left–right inconsistencies in EPI data caused by readout gradient polarity reversal. Row-wise magnitude profiles demonstrate improved consistency between odd and even lines on both sides of k-space after residual correction, corresponding to further suppression of ghost artifacts. This residual correction step provides additional improvements in image quality for both EPI and DW-EPI sequences in LF and ULF MRI, where SNR is inherently limited.

Quantitative evaluation using GSR and SNR metrics further confirms the effectiveness of the proposed method. Experiments performed on two different MRI systems demonstrate the robustness and general applicability of the algorithm, highlighting its potential for practical deployment in LF and ULF MRI.

### B. Limitations of the Correction Method

Despite its effectiveness, several limitations should be noted. The proposed reference-free approach relies on magnitude peak features extracted from row-wise k-space data. When the signal intensity is low or strongly affected by noise, peak localization may deviate from the true physical center, potentially leading to alignment errors.

In LF and ULF MRI systems, magnetic field drift may occur during relatively long acquisitions. Temperature-related instability or field fluctuations can cause odd–even line errors to exhibit non-linear behavior. Under such conditions, peak-based alignment may primarily correct the central region, while residual misalignment persists in peripheral regions, thereby reducing the effectiveness of interpolation-based residual correction.

In addition, although the proposed method effectively suppresses ghost artifacts arising from odd–even line inconsistencies, geometric distortions caused by residual k-space errors may remain. Unlike Nyquist ghost artifacts, geometric distortions can affect individual k-space lines. Ongoing work focuses on addressing these distortion effects, with preliminary results already obtained. Further improvements in distortion correction are expected to enhance EPI and DW-EPI image quality in LF and ULF MRI systems.

## Data Availability

The data used in this study were acquired from an ultra-low-field MRI prototype system. Due to the proprietary nature of the system and associated data, the datasets are not publicly available. Reasonable requests for data access may be considered by the corresponding author.

## Conflicts of Interest

The authors declare that there is no conflict of interest regarding the publication of this article.

## Funding Statement

This research did not receive any specific grant from funding agencies in the public, commercial, or not-for-profit sectors. The work was carried out as part of the authors' institutional research activities.

## Ethics Statement

All experiments involving human participants were conducted in accordance with relevant institutional guidelines and ethical standards. The volunteer experiments were performed on healthy participants, and written informed consent was obtained from all individuals prior to participation.